\def\year{2020}\relax
\documentclass[letterpaper]{article} 
\usepackage{aaai20}  
\usepackage{times}  
\usepackage{helvet} 
\usepackage{courier}  
\usepackage[hyphens]{url}  
\usepackage{graphicx} 
\usepackage{graphicx}  
\usepackage{xcolor} 
\usepackage{amsmath} 
\usepackage{amssymb} 
\usepackage{soul} 
\usepackage{booktabs} 
\usepackage{subfig} 
\title{Optical Character Recognition and Transcription of Berber Signs from Images in a Low-Resource Language Amazigh}

 \author{Levi Corallo\\
 Computational Linguistics\\
 Montclair State University\\
 Montclair, NJ, USA\\
 corallol1@montclair.edu\\
 \And
 Aparna S. Varde\\
 Computer Science Department\\
 Computational Linguistics Faculty\\
 Montclair State University, NJ, USA\\
 vardea@montclair.edu}

\begin{document}
\newcommand{\ourmodel}[1]{\textsc{CSK-SNIFFER}}
\newcommand{\yolo}[1]{\textsc{YOLO}}
 
\definecolor{Red}{rgb}{1,0,0}
\definecolor{Green}{rgb}{0,0.7,0}
\definecolor{Blue}{rgb}{0,0,1}
\definecolor{Red}{rgb}{0.6,0,0}
\definecolor{Orange}{rgb}{1,0.5,0}
\newcommand{\niket}[1]{\textcolor{Green}{[#1 \textsc{--Niket}]}}
\newcommand{\anurag}[1]{\textcolor{Blue}{[#1 \textsc{--Anurag}]}}
\newcommand{\aparna}[1]{\textcolor{Red}{[#1 \textsc{--Aparna}]}}
\newcommand{\reviewed}[1]{\textcolor{Orange}{[#1]}}

\makeatletter
\newcommand*\bigcdot{\mathpalette\bigcdot@{.5}}
\newcommand*\bigcdot@[2]{\mathbin{\vcenter{\hbox{\scalebox{#2}{$\m@th#1\bullet$}}}}}
\makeatother

\renewcommand{\v}[1]{$\mathbf{#1}$}
\newcommand{\vect}[1]{\mathbf{#1}}

\newcommand{\bluebox}[1]{\colorbox{blue!10}{#1}}
\newcommand{\redbox}[1]{\colorbox{red!10}{#1}}
\newcommand{\purplebox}[1]{\colorbox{purple!10}{#1}}



\def\DG{{\mathcal{G}}}

\newtheorem{theorem}{Definition}[section]

\newcommand{\statechange}[1]{\texttt{\textit{#1}}}
\newcommand{\entity}[1]{\texttt{#1}}
\newcommand{\strikethrough}[1]{\st{#1}}

\newcommand{\namecite}[1]{\citeauthor{#1}~\shortcite{#1}}
\newcommand{\com}[1]{}
\newcommand{\myparagraph}[1]{\vspace{1mm} \noindent {\bf #1: }}
\newcommand{\bfit}[1]{\textbf{\textit{#1}}}
\newcommand{\eat}[1]{}
\mathchardef\mhyphen="2D
\newenvironment{ite}{                     
     \parskip 0cm \begin{itemize} \parskip 0cm \parsep 0cm \itemsep 0cm \topsep 0cm}{
        \end{itemize}} 
\newenvironment{enu}{                   
     \parskip 0cm \begin{list}{}{\parsep 0cm \itemsep 0cm \topsep 0cm}}{
      \end{list}} 
\newenvironment{des}{                 
     \parskip 0cm \begin{list}{}{\parsep 0cm \itemsep 0cm \topsep 0cm}}{
      \end{list}} 
\newenvironment{myenumerate}{                   
     \parskip 0cm \begin{enumerate}{\parsep 0cm \itemsep 0cm \topsep 0cm}}{
        \end{enumerate}} 
\newenvironment{myitemize}{                     
     \parskip 0cm \begin{itemize}{\parsep 0cm \itemsep 0cm \topsep 0cm}}{
        \end{itemize}} 
\newcommand{\ourdataexpansion}{``What-If Question Answering''}
\newenvironment{myquote}{                   
  \parskip 0mm \begin{quoting}[vskip=0mm,leftmargin=2mm]}{
\end{quoting}}
\newcommand{\red}[1]{\textcolor{red}{#1}}
\newcommand{\blue}[1]{\textcolor{blue}{#1}}
\newcommand{\green}[1]{\textcolor{green}{#1}}
\newenvironment{mycentering}
 {\parskip=0pt\par\nopagebreak\centering}
 {\par\noindent\ignorespacesafterend}

    

\maketitle

\begin{abstract}
The Berber, or ``Amazigh'' language family is a low-resource North African vernacular language spoken by the indigenous Berber ethnic group. It has its own unique alphabet ``Tifinagh'' used across Berber communities in Morocco, Algeria, and others. The Afroasiatic language Berber is spoken by 14 million people, yet lacks adequate representation in education, research, web applications etc. For instance, there is no option for ``to or from Amazigh / Berber'' on Google Translate, which hosts over 100 languages today. Consequently, we do not find specialized educational apps, L2 (2nd language learner) acquisition, automated language translation, and remote-access facilities enabled in Berber. Motivated by this background, we propose a supervised approach called \textbf{DaToBS:} for \emph{ {\bf D}etection {\bf a}nd {\bf T}ranscription {\bf o}f {\bf B}erber {\bf S}igns}.  The DaToBS approach entails the automatic recognition and transcription of Tifinagh characters from signs in photographs of natural environments. This is achieved by self-creating a corpus of 1862 pre-processed character images; curating the corpus with human-guided annotation; and feeding it into an OCR model via the deployment of CNN for deep learning based on computer vision models. We deploy computer vision modeling (rather than language models) because there are pictorial symbols in this alphabet, this deployment being a novel aspect of our work. The DaToBS experimentation and analyses yield over 92\% accuracy in our pilot research. To the best of our knowledge, ours is among the first few works in the automated transcription of Berber signs from roadside images with deep learning, yielding high accuracy. This can pave the way for developing pedagogical applications in the Berber language, thereby addressing an important goal of outreach to underrepresented communities via AI in education.  
\end{abstract}


\section{Introduction}
\textbf{Motivation and Problem Definition:} 
There is much work in the literature on Optical Character Recognition (OCR), as well as language transcription and translation that spans applications in education, training, remote-access, translation and so forth. However, most of this relates to high-resource languages. These include English, German, French etc. among those with Latin scripts; Chinese (e.g. Mandarin) and Japanese (e.g. Hiragana) with pictorial scripts, and so on. Low-resource languages, i.e. ``languages or dialects that do not have resources to build technologies'' \cite{HolyGrail} often lack automated recognition, transcription and translation, for reasons such as a dearth of training corpora, lack of domain experts, and possibly less enthusiasm among the concerned people. In this paper, we focus on such a low-resource language, namely, ``Amazigh,'' scripted in its unique ``Tifinagh'' alphabet, and commonly spoken among the Berber community, constituting indigenous peoples in some parts of Northern Africa such as Morocco and Algeria. This language is spoken by around 14 million people, yet has limited resources available for education and related areas. We notice that there is a need for the following: 
\begin{itemize}
  \setlength{\itemsep}{0pt}
  \setlength{\parskip}{0pt}
    \item Optical character recognition from roadside and other outdoor images
    \item Automatic transcription from a low-resource language 
    \item Generation of a dataset that offers good pre-trained models for further learning
\end{itemize}

\begin{figure}
    \centering
    {\frame{\includegraphics[width=1.0\columnwidth]{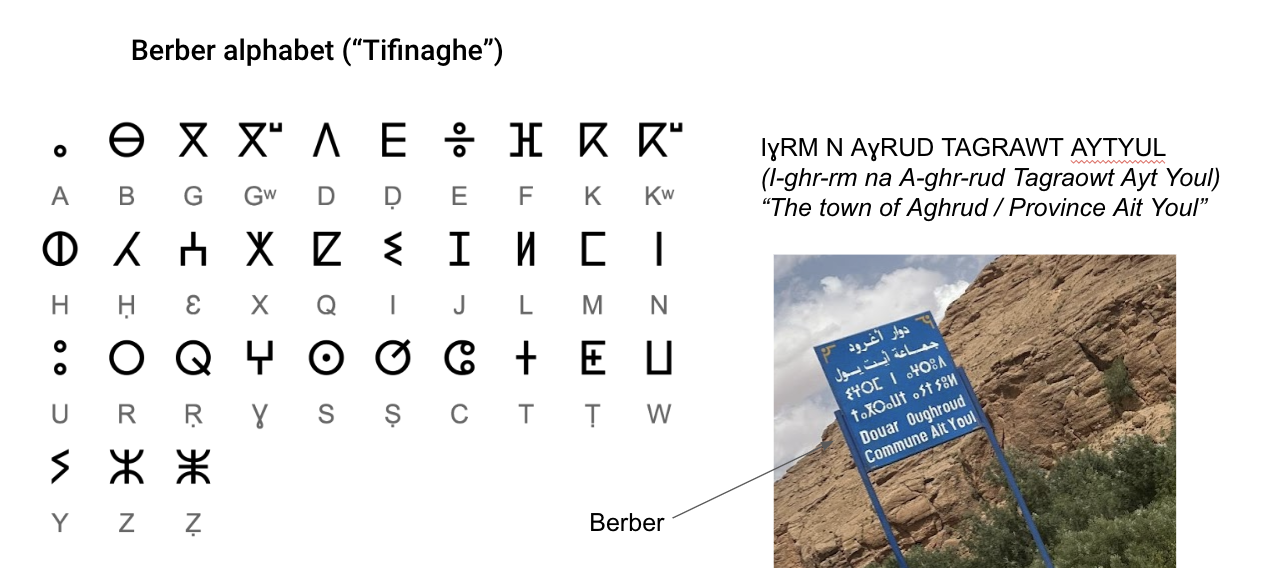}}}
    \caption{(Left) Tifinagh, the Amazigh alphabet. (Right) Trilingual road sign listing the town of Aghrud in the providence of Ait Youl, written in Arabic, Berber, and French.}
    \label{Example}
\end{figure}

Figure \ref{Example} presents a snapshot of the Berber alphabet (Tifinagh) along with a motivating example of need for language translation via an image from the roadside, depicting Berber signs.  Given this motivation, we focus our research on OCR from images that include written Berber text, and transcribe it with respect to the Tifinagh alphabet so as to make sense in the Amazigh language, while also providing a pre-trained dataset for further work such as pedagogical applications (including edutainment), camera-feeds (for teaching children, training autonomous cars etc.), and automated language translation (spanning L2 acquisition as well). Our research aims to help the masses in Berber community as well as researchers in linguistics and computer vision, by addressing NLP (Natural Language Processing) augmented by retrieved texts and linguistic features of this indigenous ethnic group. It paves the way for further research on automated translation from Amazigh into other languages such as English and French, hence providing greater recognition to low-resource language groups. It is to be noted that Google Translate currently offers language translation support for more than 100 languages, yet Amazigh / Berber / Tifinagh is not among these. Our work further sets the stage for training autonomous vehicles or travelers on foot to recognize road signs in this indigenous language to prepare for next-generation self-driving technologies across the globe. It also prepares data suitable for development of Berber language acquisition software, in both L2 and early-stage learner environments. This makes impacts on AI for education in general. Given this background and motivation, we discuss related work in the area. \\

\textbf{Related Work:} 
Machine learning research using the Berber language in OCR models has had a focus on handwriting data, with some corpora created, \cite{BencharefDataset}, and experimentation using continuous Hidden Markov Models, \cite{Handwriting}, geodesic trees, neural networks, and decision trees \cite{Handwriting2}. Methods of recognition using self-organizing maps and fuzzy K-nearest neighbor models have also been used \cite{handwriting3}, where some academic papers argue that conventional OCR methods rely too heavily on invariance and are not good enough on Tifinagh script because of how similar some characters are to one another, differing only in size or rotation. Our work hopes to provide promising accuracy leveraging deep learning well-suited for computer vision \cite{VGG}, especially considering the pictorial script. 

Similar efforts on handwriting datasets in the recent years have shown very high accuracy using deep learning \cite{Handwriting4}. The added struggle of extracting character from road signage in natural environments versus handwriting adds more potential variance to the manner in which the letters are written, as well as adding more elements that can effect a models accuracy that only appear in the wild, such as shadows, 4-dimensional lettering, and image quality/distance. Based on this brief overview of the literature, we state our own contributions in this research. \\

\textbf{Contributions:} We propose a novel approach known as \textbf{DaToBS:} \emph{ {\bf D}etection {\bf a}nd {\bf T}ranscription {\bf o}f {\bf B}erber {\bf S}igns}, which is based on supervised learning over a pictorial script. The main contributions of our work are:
\begin{enumerate}
  \setlength{\itemsep}{0pt}
  \setlength{\parskip}{0pt}
    \item Developing a self-created corpus derived from 120 natural scene images in Berber resulting in 1862 character images, curated with manually annotated bounding boxes around each character
    \item Recognizing optical characters in the Amazigh language and Tifinagh script via the adaptation of state-of-the-art object detection technology
    \item Transcribing the raw image from the natural environment to obtain the corresponding easily readable Berber text for seamless comprehension 
\end{enumerate}
We proceed to describe our DaToBS approach next, followed by its experimental evaluation and discussion. 

\section{Proposed Approach: DaToBS} 
We propose an interesting approach based on supervised learning called DaToBS, which as stated earlier is an acronym for the Detection and Transcription of Berber signs. This approach functions as its very name implies. It thrives on exploring object detection using a CNN (convolutional neural network) based architecture, more specifically computer vision modeling, with a self-created corpus consisting of Berber characters extracted from signs in natural environments. Its details are as follows. 

\textbf{Inputs:} The inputs to the approach are: 
\begin{itemize}
  \setlength{\itemsep}{0pt}
  \setlength{\parskip}{0pt}
    \item Natural scene pictures $P_S$ including self-taken photographs in various types of terrain 
    \item The Amazigh alphabet with its Tifinagh script $T$: \texttt{vocab}$(T)$ (See Figure \ref{Example} for Tifinagh script)
    \item Image search engine (e.g., Google / Bing) to query entries in \texttt{vocab}$(T)$, leading to a larger collection of images $P_T$.
\end{itemize}

\textbf{Outputs:} The outputs of the approach entail predictions of the correct label on the corresponding Berber text from the concerned image (after detection and transcription), serving the purpose of deciphering the Berber characters for comprehension in the Amazigh language.  \\

\begin{figure}
    \centering
    {\frame{\includegraphics[width=1.0\columnwidth]{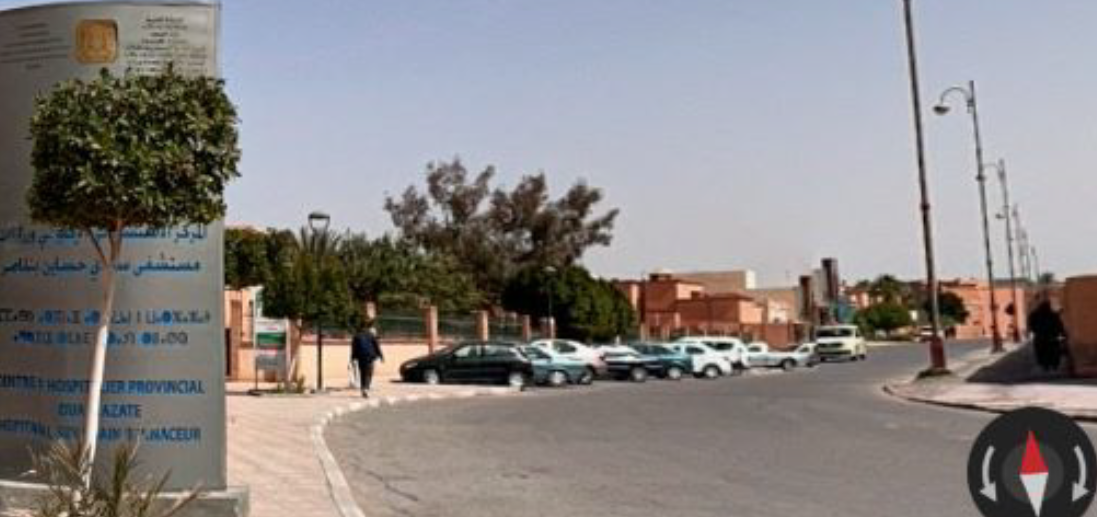}}}
    \caption{Self-taken photo from a car in Morocco with Google Street View (see compass on lower right of image)}
    \label{Photo}
\end{figure}

\textbf{Processing:} The processing in the approach entails the following. As a starting point, a good mix of images is created using the self-taken photographs $P_S$ and relevant images from Google search, $P_T$, entailing Berber street or business signage. Original photos of natural environments from a car’s viewpoint can be harvested from Google Street View (See Figure \ref{Photo}). Likewise, multiple screenshots for are taken from there to develop a self-created corpus of images. 

\begin{figure*}
  \centering
  \includegraphics[width=0.9\textwidth]{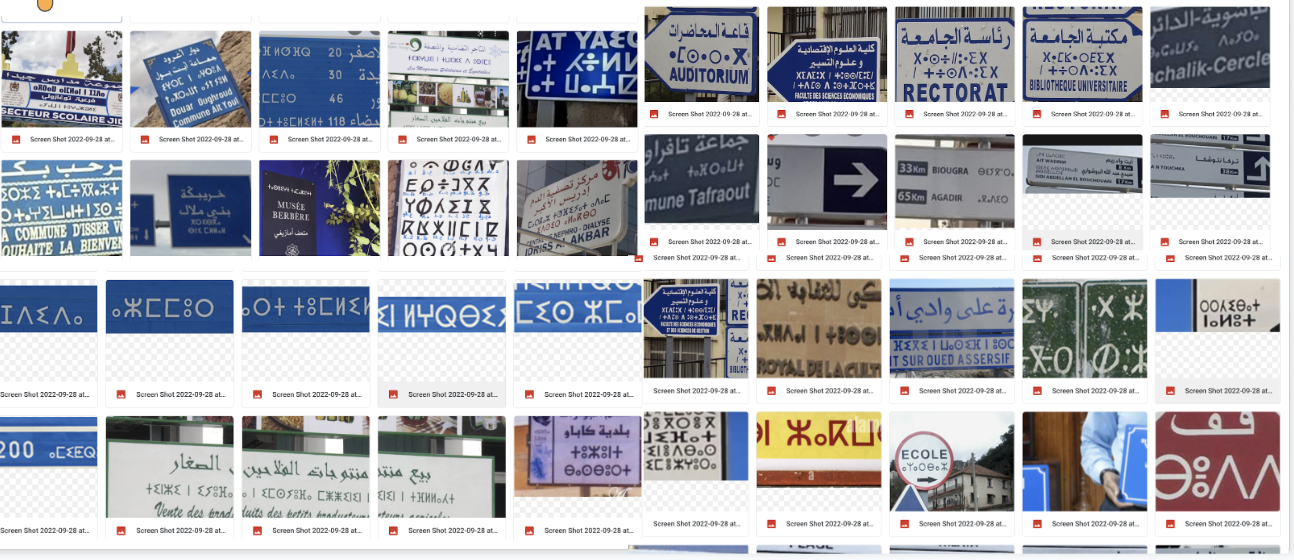}
  \caption{Corpus created for scraping on the images using domain knowledge}
  \label{figProcess}
\end{figure*}

Figure \ref{figProcess} shows the corpus before the scraping being conducted on the mix of images using knowledge about Berber and its alphabet (Tifinagh). The corpus scraping process involves manually annotating minimum-bounding boxes around characters to create new images, each containing only the Tifinagh characters from the signs (excluding other scene objects such as trees, cars etc.) that do not contain the Berber signs.  

Optimal preprocessing is then performed with a defined function that does the following:
\begin{itemize}
  \setlength{\itemsep}{0pt}
  \setlength{\parskip}{0pt}
    \item Classifies the image based on its label
    \item Iterates over each character image 
    \item Resizes the image
    \item Converts images to grey-scale
\end{itemize}

Processed images are appended to a new list that is then randomly shuffled for each execution. This is done to provide greater robustness via randomization. The images are now fed into a computer vision model VGG-16 (Visual Geometry Group - 16 layers), a convolutional neural network (CNN) based technique \cite{VGG} widely accepted in the literature for image mining \cite{yang2021novel}, \cite{karthikeyan2020transfer}, \cite{zhuang2020ifan}. Even though we deal with natural language, we prefer to deploy a computer vision model, as opposed to any PLM (pre-trained language model) such as BERT (Bidirectional Encoder Representations from Transformers) or T5 (Text-to-Text-Transfer-Transformer), since the Tifinagh script comprises pictorial characters. This is one of the novel aspects of our work. 

\begin{figure}
    \centering
    {\frame{\includegraphics[width=1.0\columnwidth]{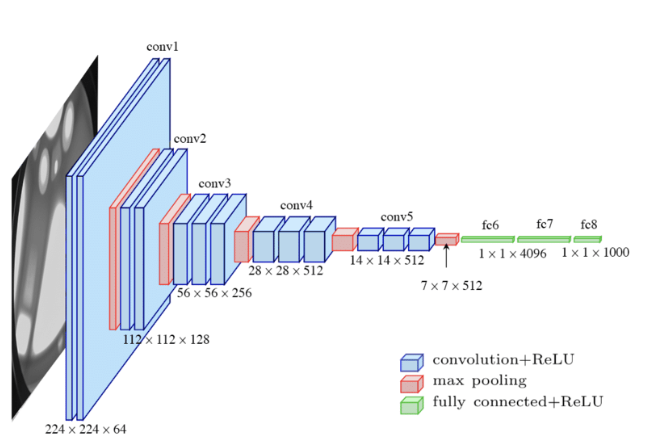}}}
    \caption{VGG-16 layout \cite{VGG}}
    \label{Arch}
\end{figure}

The VGG-16 architecture starts with one of its 13 2D convolutional layers, consisting also of 5 max-pooling layers, and 3 fully connected layers. ReLU is used as the activation function and softmax function at the output layer. VGG-16 is unique in the fact the amount of filters available start at 64, and double to 128, then 256, eventually reaching 512 filters in the final layers, as in Figure \ref{Arch}.

\begin{figure}
    \centering
    {\frame{\includegraphics[width=1.0\columnwidth]{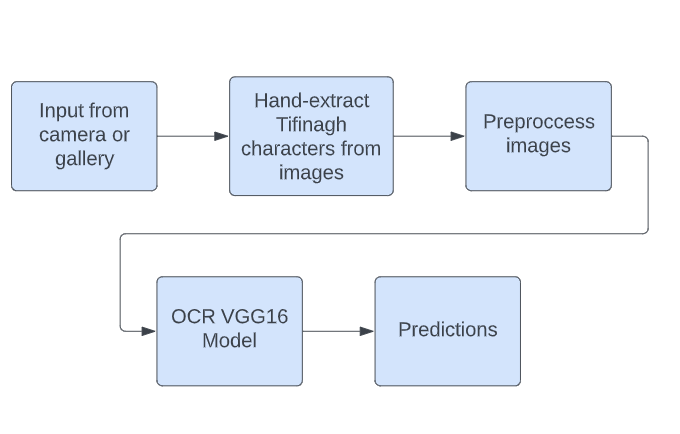}}}
    \caption{The flow of data from creation to prediction}
    \label{flow}
\end{figure}

The training split of images is then used to train the VGG-16 model, where an accuracy is given for performance on both the training and test sets. Loss, accuracy, and confusion matrix charts are generated based on the data from the execution that help draw further analyses onto which characters caused incorrect predictions. The entire data flow in the processing of DaToBS is illustrated in Figure \ref{flow}. We now discuss our experiments.    

\section{Experimental Evaluation}
\textbf{Experimental Setup:} 
In our experiments, 120 images, entailing a combination of self-taken photographs and related images from a Google search, (as stated in the inputs of DaToBS) are hand-scraped to extract the characters from photographs containing Berber road signs, business signage, and other natural environment photographs with readable text on signs. The resulting 1862 character images are then placed into each of the corresponding 33 character folders in order to conduct the supervised training. A training/test split of 80\%/20\% is created. OpenCV \cite{opencv} is used to standardize character images to a uniform 50x50px black and white version of the character image. Processed images are serialized with Pickle \cite{pickle}, and reshaped and prepared to be fed into the VGG-16 model. The model initialization uses Keras \cite{keras}, where the number of classes is set to 33 for each of the respective Tifinagh characters. Learning rate of 0.001 is chosen, after multiple trials, because we find that it provides the best results on this particular dataset. Batch-size and epochs are also experimented upon, among which a few executions are summarized here. \\

\textbf{Observation on Character Occurrence Imbalance}:
It is interestingly noted that some Tifinagh characters appear much more frequently than others in roadside signs. For example, the top three largest character images constitute for over 1/3 of the entire character image corpus (34.3\%). Those top three occurring characters are the respective Berber letters for ``a,'' ``n,'' and ``i.'' The disproportionality reflects the natural occurrence of some characters over others on actual signage in Moroccan towns and other Berber-fluent communities. While a couple of characters dominate, some others are almost never seen printed on a road sign because they are produced as vernacular emphatic versions of the Tifinagh characters e.g. ``gh'' versus ``ghw'' which differ by being marked with a diacritical symbol, signaling a difference in pronunciation. The occurrence of the emphatic ``zz'' was 0.37\% in our corpus. Such observations can be crucial, especially in pedagogy, and can guide further work. 

\begin{table}[ht]
    \centering
    \begin{tabular}{|c|c|} \hline
    Execution     & Accuracy\\ 
    \hline
    1: 16batch-size, 12epochs & 95.3\%train / 89.4\%test \\  
    2: 18batch-size, 10epochs  & 91.2\%train / 88.9\%test \\
    3: 15batch-size, 12epochs & 94.4\%train / 92.2\%test\\
    \hline

    \end{tabular}
    \caption{Accuracy of DaToBS in evaluation results}
    \label{tab:resu16lts}
\end{table}

\begin{figure}
    \centering
    {\frame{\includegraphics[width=1.0\columnwidth]{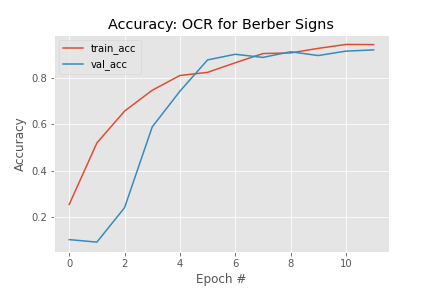}}}
    \caption{Accuracy per epoch on Execution 3 that achieves 92.2\% accuracy overall}
    \label{AccuracyPlot}
\end{figure}

\begin{figure}
    \centering
    {\frame{\includegraphics[width=1.0\columnwidth]{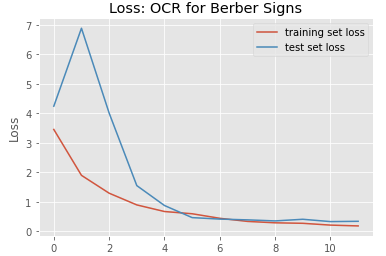}}}
    \caption{Loss per epoch on Execution 3 that achieves 92.2\% accuracy overall}
    \label{LossPlot}
\end{figure}

\noindent\textbf{Experimental Results:} 
Interesting results from our experimentation are tabulated here, depicting three significant executions, to observe how the VGG-16 model handles the corpus of Tifinagh characters web-scraped from photographs. The overall results are synopsized in Table 1 while plots depicting accuracy and loss for the Execution 3 (in this paper) appear in Figure \ref{AccuracyPlot} and \ref{LossPlot} respectively. We find that our results for an OCR model trained on images from natural environments is realistic, and is comparable to research on OCR for Tifinagh handwriting which has much less invariance and noise added due to image source data being from the wild. Through such experimentation, we anticipate expanding the ongoing field of machine learning on Tifinagh characters to serve the Berber / Amazigh population. In addition to the impact of providing pre-trained datasets to pave the way for automated translation from the Berber language, this anticipation comes with the promising initiative of next-generation technology, e.g. self-driving cars, and pedagogical apps in countries such as Morocco. It envisages providing modern technology with the ability to seamlessly compute road-signs in an automated manner as a computer vision model runs within a vehicle's camera, as well as developing relevant apps for educational and training purposes. In this discussion, we wish to add a note that our code and annotated dataset can be made publicly available on GitHub. We are not including it in the paper due to anonymity for double-blind review.

\section{Discussion and Educational Impacts}
\noindent\textbf{Benchmarking v. Handwriting OCR with Amazigh:} Berber's unique 33 character set making up its alphabet Tifinagh is observed to contain many possible obstructions in our corpus on the characters that make the task of OCR from natural environments harder than handwriting for OCR, which is an important point to note from educational perspectives. Our data-collection and analyses indicate that the claims stand true for the Berber language as well, posing additional challenges. Data from our Amazigh signage corpus finds instances of raw image data where characters are:
\begin{itemize}
  \setlength{\itemsep}{0pt}
  \setlength{\parskip}{0pt}
    \item varied in the style they occurred on the sign e.g. standard road marking paint, regular paint, characters in 3-D casting shadows on signs, typeface, etc. 
    \item varied in the distance they were from the input camera source which affected their size and image-quality
    \item  obstructed by other objects outdoors
    \item naturally eroded
    \item varied in the background of the character
\end{itemize}
These variables which must account for potential change in the data harvested from Berber natural scenes are not found in The Amazigh Handwritten Character Database (AMHCD), which is the only large database of handwritten Berber characters \cite{HandwritingDS}. AMHCD, developed at the IRF-SIC Laboratory of Agadir, Morocco, is used to benchmark results. Our VGG-16 model with its best parameters is fed with the Amazigh handwritten character corpus. Each character folder contains 780 handwritten characters curated from 60 different writers, for a total of 25,740 images. Experimentation is conducted in order to ground the claim that OCR in natural environments is a harder task versus handwriting OCR in Berber as well. A 80\%/20\% train/test set is also used. Results confirm our claims, where the same VGG-16 parameters with the AMHCD dataset achieves 98\%+ test-set accuracy, 6\% higher than the best results our natural environment dataset. \\

\noindent\textbf{Benchmarking with Another Model:} In order to offer a comprehensive analysis benchmarking VGG-16's accuracy, a different yet similar model, VGG-19 is experimented with using the DaToBS corpus. This raises an interesting and rather surprising point: since the very sparse amount of Berber image data available to be web-scrapped alongside the very little amount of data resources available for Berber, the size of DaToBS is forced to be limited. Given that the corpus size is considered very low yet suitable for deep learning, the 16-layer architecture of VGG-16 is found to be preferable when dealing with OCR datasets of this size. When using VGG-19, we find that accuracy values plummet with results failing to break 20\%+ on the same data. Thus, we find VGG-16 to be preferable, as per the observations in our pilot studies. \\

\begin{figure}
    \centering
    {\frame{\includegraphics[width=1.0\columnwidth]{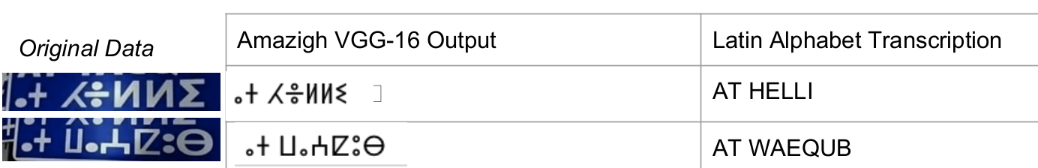}}}
    \caption{Sample outputs from DaToBS for use in a parallel corpus, helpful for future tasks in machine translation}
    \label{llmtable}
\end{figure}

\noindent\textbf{Benchmarking with LLMs:} To test whether OCR accuracies with the DaToBS corpus could be improved pairing word-level output with large language models, the Tesseract library \cite{Tesseract} was used to generate comparative results with this RNN (Recurrent Neural Network) LSTM (Long Short Term Memory) model. Tesseract provides PLMs (pre-trained language models) for over 100 languages, yet again with Amazigh being absent. Accuracies were noted to raise 6\%+, where Arabic words were detected with 98\%+ accuracy, where 100 out of 102 instances of Arabic words were correctly identified from the DaToBS corpus using Tesseract's Arabic PLM. This finding further incentives the need for a Amazigh LLM, to be used in OCR models like this one for the Berber community, but also getting this endangered alphabet up to speed in both the current era of rapidly accelerating  NLP progress. It would also be an effort towards preserving and documenting natural language taken from Amazigh/Tamazight speakers and transcribed in the Tifinagh character set alphabet. In Fig.  \ref{llmtable}, the outputs from the DaToBS corpus across experimentation can even act as a beginning effort towards a Berber LLM fine-tuned for geographical or scenic text recognition NLP tasks.

\begin{figure}
    \centering
    {\frame{\includegraphics[width=1.0\columnwidth]{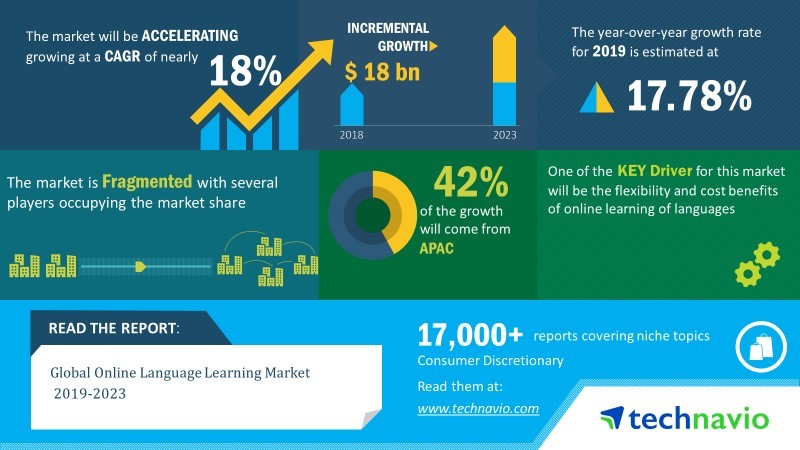}}}
    \caption{Information from Technavio on the booming language acquisition software market the disproportionate focus on certain languages}
    \label{Photo7}
\end{figure}

\begin{figure}
    \centering
    {\frame{\includegraphics[width=0.8\columnwidth]{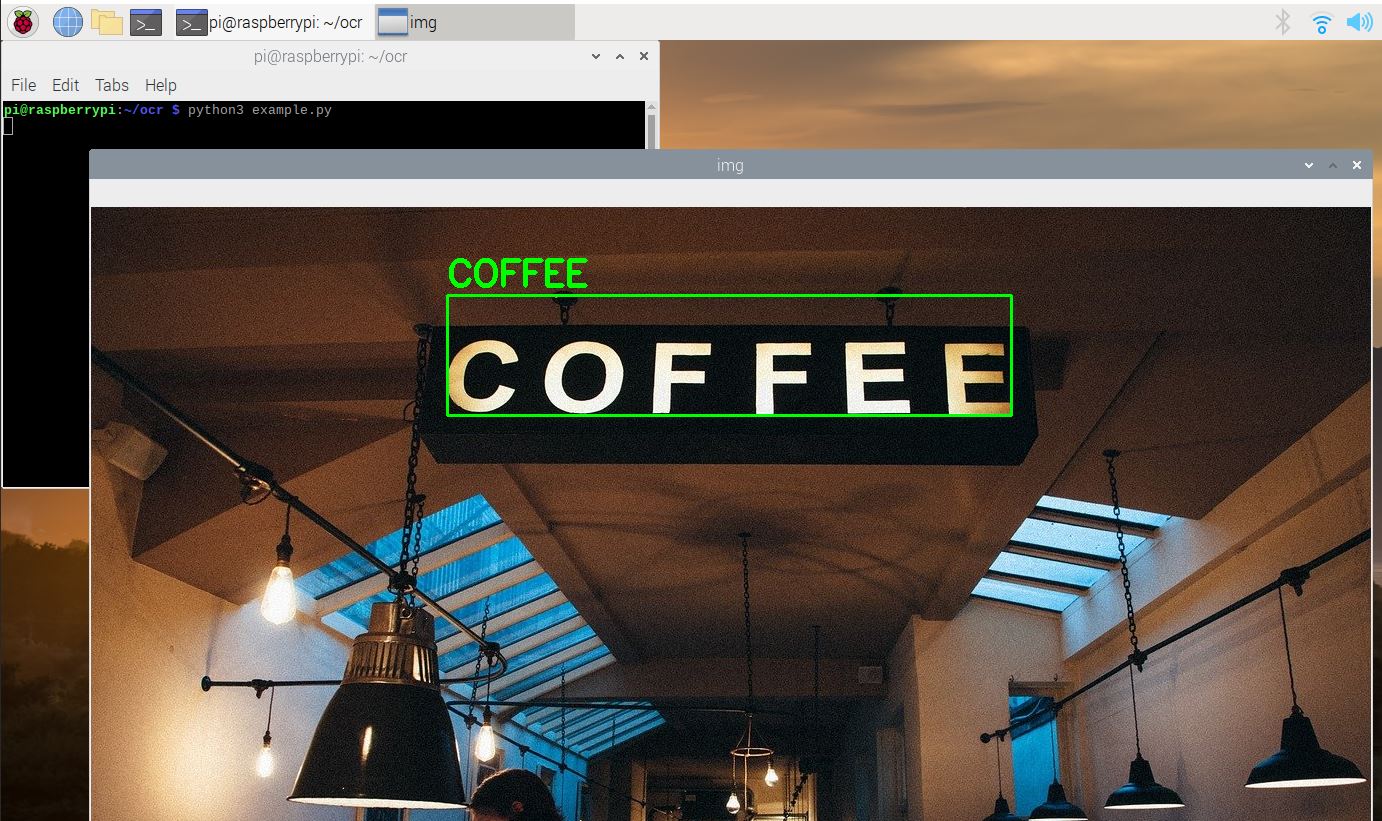}}}
    \caption{An example of live-camera feed text transcription}
    \label{Photo8}
\end{figure}

\begin{figure}
    \centering
    {\frame{\includegraphics[width=0.8\columnwidth]{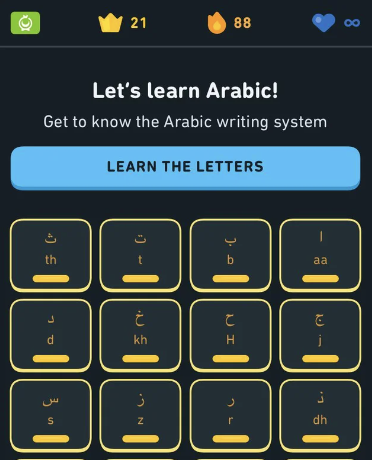}}}
    \caption{A screenshot from Duolingo Arabic language acquisition phone application software (applications analogous to this can be designed for Berber in Amazigh)}
    \label{Photo9}
\end{figure}

\begin{figure}
    \centering
    {\frame{\includegraphics[width=0.5\columnwidth]{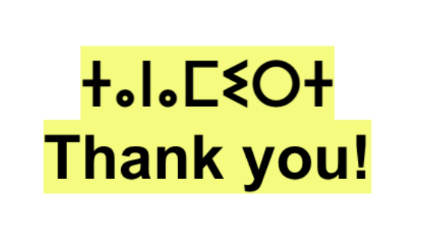}}}
    \caption{Thank you in Berber, pronounced ``tanmirt''}
    \label{tanmirt}
\end{figure}

\noindent\textbf{Impacts on 21st Century Education:}
The DaToBS model and corpus are the first of its kind for OCR from raw environment data within the Berber language. The work in this paper hopes to be able to help contribute to the low-resource Berber NLP research collective, and also aims to provide potential educational benefits as the model gets deployed. DaToBS can pave the way for developing the following. 
\begin{itemize}
  \setlength{\itemsep}{0pt}
  \setlength{\parskip}{0pt}
    \item Berber L2 language acquisition software, which does not yet exist (see Fig. \ref{Photo7})
    \item Software for camera-feed identification of the alphabet e.g. for early-stage learner children in Berber-speaking households, or for travelers and autonomous cars to read road signs (see Fig. \ref{Photo8})
    \item Other pedagogical applications that can provide education and training to the masses on the richness, intricacies and nuances of the Berber language, thereby contributing to low-resource languages (see Fig. \ref{Photo9})
\end{itemize}
While the amount of models and resources available for Berber are still lacking, (e.g. no large language model exists for Berber) our work via its developed corpus aims to contribute to the amount of Amazigh language resources that are available. This research focus also hopes to enable more research-level educational tasks in machine learning for interested Berber-fluent residents in countries like Morocco or Algeria, thus constituting additional impacts of the work. Fig. \ref{tanmirt} shows an early phrase that would be apart of L2 mobile Berber acquisition. This OCR model specifically could pave the way for Berber judging L2 mobile learner's correctness to draw characters with touch-screen on smartphones. 

DaToBS is in line with some of our earlier research impacting 21st century education, e.g. \cite{Corallo2022} on German-English news translation, \cite{Bhagat2019} on preposition prediction in written English, and \cite{Varghese2015} on correction of odd collocations for L2 English (second language learners). It can also make broader impacts on smart governance in general and app development in particular, augmenting prior work by our research group, e.g. \cite{Puri2018}, \cite{Basavaraju2016}. Moreover, it can contribute to the extraction and compilation of commonsense knowledge which is of much interest today as evident from recent tutorials \cite{Tandon2017}, \cite{Razniewski2021}. Further to such work, cultural commonsense is vital and some culture-specific knowledge can be extracted from local-language text within images. All these constitute modest impacts of our work on AI in education. 

\section{Conclusions and Future Work}
We propose an approach called DaToBS for the detection and transcription of Berber signs, thus making modest contributions vis-a-vis a low-resource language Amazigh. We find that harnessing a deep learning computer vision model VGG-16 (derived from a CNN architecture) is effective enough for a pilot study in performing OCR on the Berber language so as to be potentially useful in language translation as well as live camera-feed technology. This corpus and model hopes to also enable future experimentation by open-sourcing the annotated dataset, as well as be resourceful to future projects on Berber language acquisition, and as an educational tool for travelers. Future corpora development for Tifinagh characters as well as datasets to work towards Berber / Amazigh large language models are much needed in this field as its alphabet is considered endangered despite its large population of speakers (14million+). Overall, achieving 92\%+ in the DaToBS approach at an initial stage is a good indicator of success, proving specifically effective in OCR for Berber characters. Looking forward, we can focus on the following aspects of future work. 
\begin{itemize}
  \setlength{\itemsep}{0pt}
  \setlength{\parskip}{0pt}
    \item Continue to expand the DaToBS corpus with raw image data containing Berber characters 
    \item Conduct experimentation using other CNN based computer vision models, e.g. ResNet-101
    \item Perform comparative studies with PLMs such as BERT and T5, in order to further gauge how language models work in the concerned tasks
    \item Determine whether significantly more data samples will create a proportionately more robust model that could be beneficial for raw images or live camera-feeds as well as test whether character imbalances are still reflected
    \item Annotate an expanded version of the natural images with bounding boxes at word-level instead of just character-level to create a more powerful, adaptable model 
    \item Consider an approach to train the model using the raw images rather than the single-character images and try to recognize multiple characters on one image simultaneously.
\end{itemize}
This work contributes two cents to the field of AI in education, by augmenting existing knowledge with retrieved texts and linguistic features of a low-resource language Amazigh, and consequently helping to set the stage for building more advanced applications. It thus aims to offer more recognition to this language and the respective indigenous peoples, which is an important aspect of outreach with respect to educational initiatives worldwide. It provides a reasonably sized self-generated, curated and pre-trained dataset usable by other researchers. Finally, we anticipate that it can encourage much further work on Berber language acquisition, language translation, pedagogical apps, and potential software using live camera-feeds on the Berber language, in line with next generation technology for greater automation within language education.

\section{Acknowledgements}
Levi Corallo would like to extend his thanks to Montclair State University, where this research was completed and who have also sponsored his registration for AI4ED at AAAI-23. He also gives special thanks to Jaouad Ait-Haddou, who was consulted as a native Berber speaker throughout the pilot research, as well as all of his family and friends who have been supportive throughout the process.  Aparna Varde acknowledges NSF grants 2117308 (MRI: Acquisition of a Multimodal Collaborative Robot System (MCROS) to Support Cross-Disciplinary Human-Centered Research and Education at Montclair State University), and 2018575 (MRI: Acquisition of a High-Performance GPU Cluster for Research and Education). She is a visiting researcher at the Max Planck Institute for Informatics in Saarbrucken, Germany, ongoing from her sabbatical, conducting data science related research, spanning NLP and commonsense knowledge.

\bibliography{references}
\bibliographystyle{aaai}

\end{document}